\pdfoutput=1

\documentclass[11pt]{article}

\usepackage{acl}

\usepackage{times}
\usepackage{latexsym}
\usepackage{amsmath}
\usepackage{amsfonts}
\usepackage{bm}
\usepackage{url}
\usepackage{graphicx}
\usepackage{multicol}
\usepackage{multirow}
\usepackage{amssymb}
\usepackage{booktabs}
\usepackage[T1]{fontenc}

\usepackage[utf8]{inputenc}

\usepackage{microtype}
\usepackage{hyperref}
\usepackage{xspace}
\usepackage{paralist}
\usepackage{booktabs}
\usepackage{mdwlist}
\usepackage{enumitem}
\usepackage{color}
\usepackage{dsfont}
\usepackage{amsthm}
\usepackage{pifont}
\usepackage{bm}
\usepackage{subfig}
\usepackage{algorithm,algpseudocode}

\newcommand{\method}{STEMM\xspace}

\definecolor{MyGreen}{HTML}{2BA02D}
\definecolor{MyOrange}{HTML}{FF7F0E}
\definecolor{MyBlue}{HTML}{1F77B4}

\usepackage{algorithm}
\usepackage{algpseudocode}

\title{\method: Self-learning with Speech-text Manifold Mixup for Speech Translation}

\author{
    Qingkai Fang\textsuperscript{\rm 1,2\textdagger},
    Rong Ye\textsuperscript{\rm 3},
    Lei Li\textsuperscript{{\rm 4}*\textdagger},
    Yang Feng\textsuperscript{{\rm 1,2}*},
    Mingxuan Wang\textsuperscript{{\rm 3}*} \\
    \textsuperscript{\rm1} Key Laboratory of Intelligent Information Processing \\ Institute of Computing Technology, Chinese Academy of Sciences (ICT/CAS) \\
    \textsuperscript{\rm2} University of Chinese Academy of Sciences, Beijing, China \\
    \textsuperscript{\rm3} ByteDance AI Lab \quad \textsuperscript{\rm4} University of California, Santa Barbara \\
    \texttt{\{fangqingkai21b,fengyang\}@ict.ac.cn} \\
    \texttt{\{yerong,wangmingxuan.89\}@bytedance.com,}\quad \texttt{leili@cs.ucsb.edu}
}

\usepackage{lipsum}
\newcommand\blfootnote[1]{%
  \begingroup
  \renewcommand\thefootnote{}\footnote{#1}%
  \addtocounter{footnote}{-1}%
  \endgroup
}

\begin{document}
\maketitle
\blfootnote{* indicates corresponding authors.}
\blfootnote{\textsuperscript{\textdagger} Work was done while at ByteDance AI Lab.}
\blfootnote{Part of joint project between ICT/CAS and ByteDance AI Lab. Work was done when QF was a member of the joint project. }
\blfootnote{Code and models are publicly available at \url{https://github.com/ictnlp/STEMM}.}

\begin{abstract}
How to learn a better speech representation for end-to-end speech-to-text translation (ST) with limited labeled data? Existing techniques often attempt to transfer powerful machine translation (MT) capabilities to ST, but neglect the representation discrepancy across modalities. In this paper, we propose the \textbf{S}peech-\textbf{TE}xt \textbf{M}anifold \textbf{M}ixup (\method) method to calibrate such discrepancy. Specifically, we mix up the representation sequences of different modalities, and take both unimodal speech sequences and multimodal mixed sequences as input to the translation model in parallel, and regularize their output predictions with a self-learning framework. Experiments on MuST-C speech translation benchmark and further analysis show that our method effectively alleviates the cross-modal representation discrepancy, and achieves significant improvements over a strong baseline on eight translation directions. 
\end{abstract}

\section{Introduction}
\label{sec:intro}
Speech-to-text translation (ST) aims at translating acoustic speech signals into text in a foreign language, which has wide applications including voice assistants, translation for multinational video conferences, and so on.
Traditional ST methods usually combine automatic speech recognition (ASR) and machine translation (MT) in a cascaded manner~\citep{sperber-etal-2017-neural, cheng-etal-2018-towards, sperber-etal-2019-self, Dong_Wang_Yang_Chen_Xu_Xu_2019, zhang-etal-2019-lattice, lam2021cascaded}, which might suffer from error propagation and high latency. 
To break this bottleneck, end-to-end ST systems attracted much attention recently \citep{wang2020bridging, wang-etal-2020-curriculum, dong2021consecutive, dong2021listen, han-etal-2021-learning, inaguma-etal-2021-source, tang-etal-2021-improving}, which learn a unified model to generate translations from speech directly. Some recent work has shown great potential for end-to-end speech translation, even surpassing traditional cascaded systems \citep{ye2021end, xu-etal-2021-stacked}.

\begin{figure}[t]
    \centering
    \includegraphics[width=\linewidth]{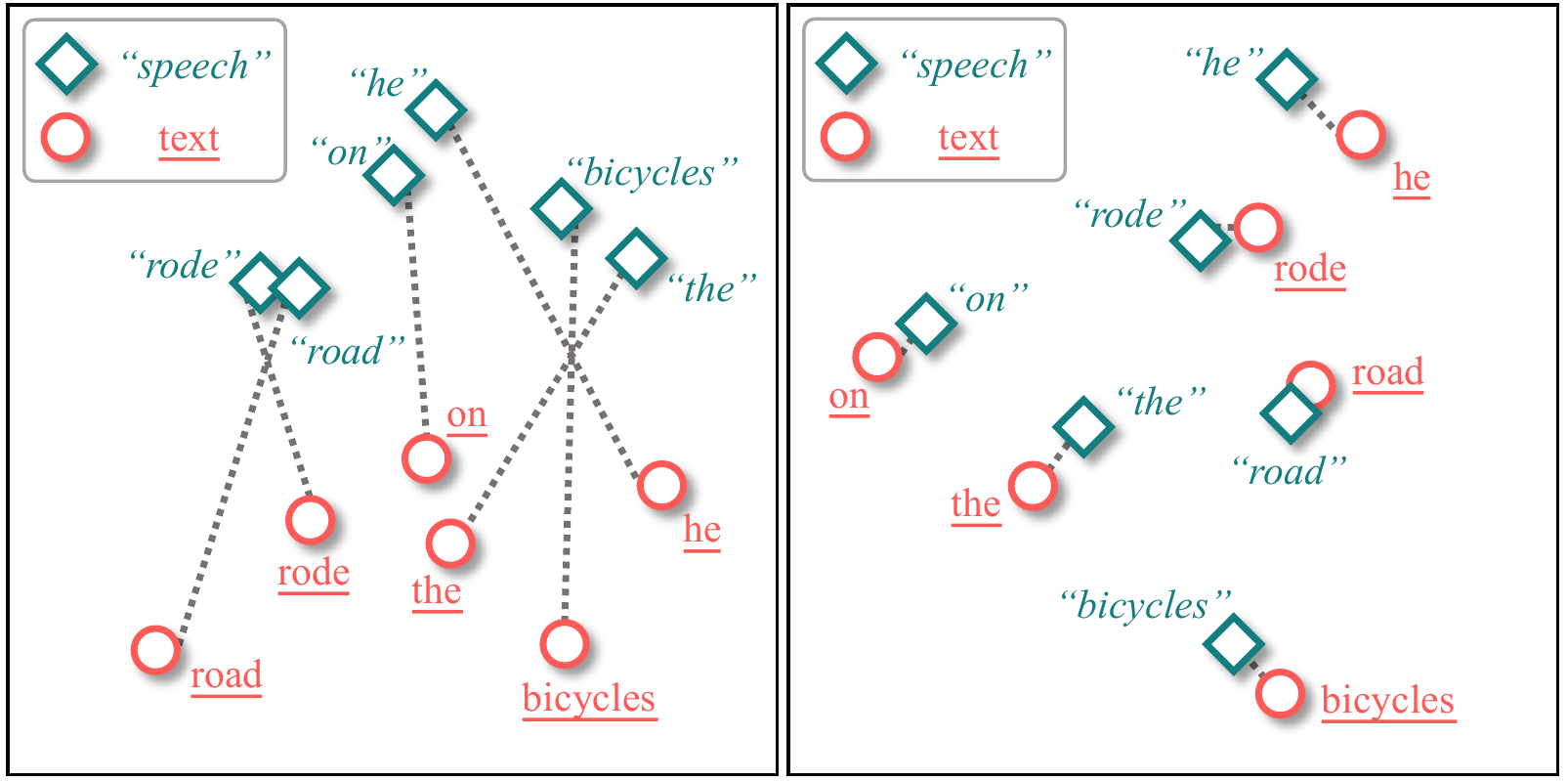}
   \caption{\method aims at bridging the modality gap of speech and text.   Different modalities with the same meaning are projected to a shared space.}
    \label{fig:expected}
\end{figure}

As a cross-modal task, a major challenge in training an end-to-end ST model is the representation discrepancy across modalities, which means there is a modality gap between speech representations and text embeddings, as shown in the left sub-figure of Figure~\ref{fig:expected}.
Existing approaches often adopt a sophisticated MT model to help the training of ST, with some techniques like pretraining~\citep{wang-etal-2020-curriculum,ye2021end, xu-etal-2021-stacked}, multi-task learning~\citep{ye2021end,han-etal-2021-learning, tang-etal-2021-improving} and knowledge distillation~\citep{liu2019end, gaido2020end,inaguma2021source, tang-etal-2021-improving}. Although these methods have achieved impressive improvements in ST task, these methods are not necessarily the best way to leverage the MT knowledge.
Considering that during training, the input of the translation module only include speech sequences or text sequences, the lack of multimodal contexts makes it difficult for the ST model to learn from the MT model.
Inspired by recent studies on some cross-lingual \citep{lample2019cross, liu2020multilingual, lin2020pre} and cross-modal \citep{li2021unimo, zhou2020unified, dong2019unified} tasks, we 
suggest that  building a shared semantic space between speech and text, as illustrated in the right sub-figure of Figure~\ref{fig:expected},  has the potential to benefit the most from the MT model.


In this paper, we propose the \textbf{S}peech-\textbf{TE}xt \textbf{M}anifold \textbf{M}ixup (\method) method to bridge the modality gap between text and speech. In order to calibrate the cross-modal representation discrepancy, we mix up the speech  and text representation as the input and keep   the target sequence unchanged. 
Specifically, \method is a self-learning framework, which takes both the speech representation and the mixed representation as parallel inputs to the translation model, and regularizes their output predictions. 
Experimental results show that our method achieves promising performance on the benchmark dataset MuST-C \citep{di-gangi-etal-2019-must}, and even outperforms a strong cascaded baseline. Furthermore, we found that our \method could effectively alleviate the cross-modal representation discrepancy, and project two modalities into a shared space.

\section{Method}

\begin{figure}[t]
    \centering
    \includegraphics[width=\linewidth]{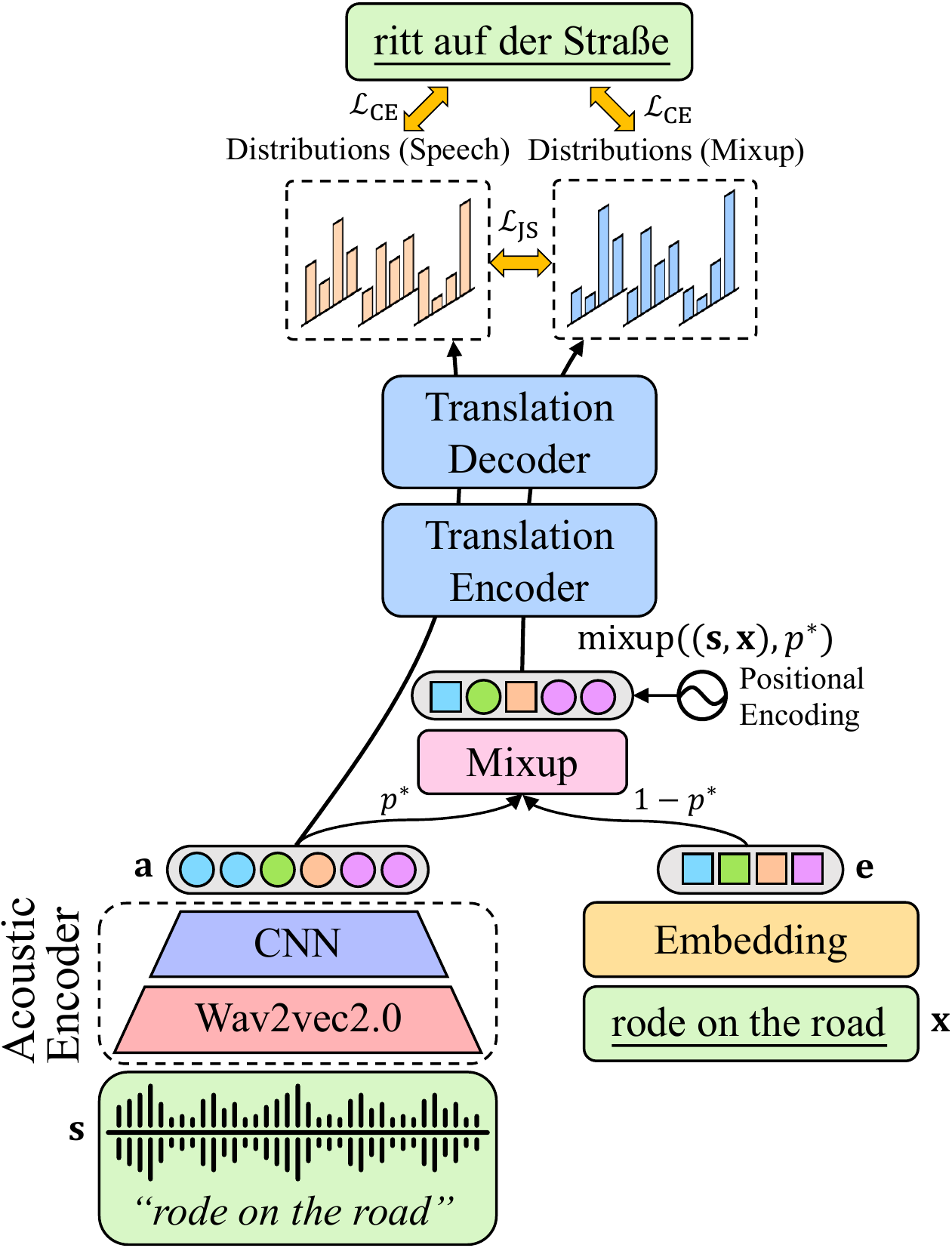}
    \caption{Overview of our proposed self-learning framework with \method. We first mix up the sequence of speech representations and word embeddings with \method. Then, both the unimodal speech sequence and the multimodal mixed sequence are fed into the shared translation module to predict the translation, and we regularize two output predictions with an additional JS Divergence loss.}
    \label{fig:method}
\end{figure}

In this section, we will begin with the basic problem formulation (Section \ref{sec:problem-formulation}) and introduce the model architecture (Section \ref{sec:model}). Then, we introduce our proposed \textbf{S}peech-\textbf{TE}xt \textbf{M}anifold \textbf{M}ixup (\method) in Section \ref{sec:mixup}. Finally, we introduce our proposed self-learning framework with \method in Section \ref{sec:self-training} and present two mixup ratio strategies in Section \ref{sec:mixup-strategy}. Figure \ref{fig:method} illustrates the overview of our proposed method.

\subsection{Problem Formulation}
\label{sec:problem-formulation}
The speech translation corpus usually contains \emph{speech-transcription-translation} triples, which can be denoted as $\mathcal{D}=\{(\mathbf{s}, \mathbf{x}, \mathbf{y})\}$. Here $\mathbf{s}$ is the sequence of audio wave, $\mathbf{x}$ is the transcription in the source language, and $\mathbf{y}$ is the translation in the target language. End-to-end speech translation aims to generate translation $\mathbf{y}$ directly from the audio wave $\mathbf{s}$, without generating intermediate transcription $\mathbf{x}$.

\subsection{Model Architecture}
\label{sec:model}
Inspired by recent works \citep{dong2021listen, xu-etal-2021-stacked} in end-to-end speech translation, we decompose the ST model into three modules: \emph{acoustic encoder}, \emph{translation encoder}, and \emph{translation decoder}. The \emph{acoustic encoder} first encodes the original audio wave into hidden states, fed into the \emph{translation encoder} to learn further semantic information. Finally, the \emph{translation decoder} generates the translation based on the output of the \emph{translation encoder}. 

\noindent\textbf{Acoustic Encoder}~
As recent works~\citep{ye2021end, han-etal-2021-learning} show that Wav2vec2.0~\citep{baevski2020wav2vec} can improve the performance of speech translation, we first use a pretrained Wav2vec2.0 to extract speech representations $\mathbf{c}$ from the audio wave $\mathbf{s}$. 
We add two additional convolutional layers to further shrink the length of speech representations by a factor of 4, denoted as $\mathbf{a}=\text{CNN}(\mathbf{c})$. 

\noindent\textbf{Translation Encoder}~
Our \emph{translation encoder} is composed of $N_e$ transformer \citep{transformer} encoder layers, which includes a self-attention layer, a feed-forward layer, normalization layers, and residual connections. 
For MT task, the input of the \emph{translation encoder} is the embedding of transcription $\mathbf{e}=\text{Emb}(\mathbf{x})$. For ST task, it is the output sequence of the \emph{acoustic encoder} $\mathbf{a}$. The input can also be the multimodal mixed sequence with our proposed \method (see details in Section \ref{sec:mixup}). 
Generally, for the input sequence $\bm{\chi}$, we obtain the contextual representations $\mathbf{h}(\bm{\chi})$ after $N_e$ transformer \citep{transformer} layers, which are fed into the \emph{translation decoder} for predicting the translation. 

\noindent\textbf{Translation Decoder}~
Our \emph{translation decoder} is composed of $N_t$ transformer decoder layers, which contain an additional cross-attention layer compared with transformer encoder layers. For the input sequence $\bm{\chi}$, the cross entropy loss is defined as:
\begin{equation}
    \label{eq:trans}
    \mathcal{L}_{\text{CE}}(\bm{\chi}, \mathbf{y}) = -\sum_{i=1}^{|\mathbf{y}|}\log p_\theta(\mathbf{y}_i|\mathbf{y}_{<i}, \mathbf{h}(\bm{\chi})).
\end{equation}

\noindent\textbf{Pretrain-finetune}~
We follow the pretrain-finetune paradigm to train our model. First, we pretrain the \emph{translation encoder} and \emph{translation decoder} with parallel \emph{transcription-translation} pairs, derived from both the speech translation corpus and the external MT dataset. Also, the \emph{acoustic encoder} is pretrained on large amounts of unlabeled audio data in a self-supervised manner. We combine those pretrained modules and finetune the whole model for ST. 

\subsection{Speech-Text Manifold Mixup (\method)}

\label{sec:mixup}

As we mentioned in Section \ref{sec:intro}, to alleviate the representation discrepancy due to the lack of multimodal contexts, we present the \textbf{S}peech-\textbf{TE}xt \textbf{M}anifold \textbf{M}ixup (\method) method to mix up the sequence of speech representations and word embeddings. We first introduce \method in this section and later show how to use it to help the training of ST.

Note the sequence of sub-word embeddings as $\mathbf{e}=[\mathbf{e}_1, \mathbf{e}_2, ..., \mathbf{e}_{|\mathbf{e}|}]$ and the sequence of speech representations as $\mathbf{a}=[\mathbf{a}_1, \mathbf{a}_2, ..., \mathbf{a}_{|\mathbf{a}|}]$, where the sequence lengths usually follow $|\mathbf{a}| \geq |\mathbf{e}|$.
We first perform a word-level forced alignment between speech and text transcriptions to determine when particular words appear in the speech segment.
Formally, the aligner recognizes a sequence of word units $\mathbf{w}=[w_1, w_2, ..., w_T]$, and for each word $w_i$, it returns the start position $l_i$ and end position $r_i$ in the sequence of speech representation $\mathbf{a}$.
Meanwhile, we denote the corresponding sub-word span for word $w_i$ as $[x_{m_i}: x_{n_i}]$, with its embeddings matrix $[\mathbf{e}_{m_i} : \mathbf{e}_{n_i}]$, where $m_i$ and $n_i$ are the start position and end position in the sequence of sub-words.
To mix up both sequences, for each word unit $w_i$, we choose either the segment of speech representations $[\mathbf{a}_{l_i}:\mathbf{a}_{r_i}]$ or sub-word embeddings $[\mathbf{e}_{m_i} : \mathbf{e}_{n_i}]$ with a certain probability $p^*$, referred to \emph{mixup ratio} in this paper.

\begin{equation}
    \mathbf{m}_i=\begin{cases}
        [\mathbf{a}_{l_i}:\mathbf{a}_{r_i}] & p \leq p^* \\
        [\mathbf{e}_{m_i}:\mathbf{e}_{n_i}] & p > p^*
    \end{cases},
\end{equation}
where $p$ is sampled from the uniform distribution $\mathcal{U}(0, 1)$.


Finally, we concatenate all $\mathbf{m}_i$ together and obtain the mixup sequence:
\begin{equation}
    \mathbf{m} = \text{Concat}(\mathbf{m}_1, \mathbf{m}_2, ..., \mathbf{m}_T).
\end{equation}

Note that in terms of the mixup representation sequence length, we have $|\mathbf{e}| \leq |\mathbf{m}| \leq |\mathbf{a}|$. 
Considering the positions of tokens have changed after mixup, we add positional encodings to the token embeddings. We further perform layer normalization to normalize the embeddings:
\begin{equation}
    \text{Mixup}((\mathbf{s}, \mathbf{x}), p^*) = \text{LayerNorm}(\mathbf{m} + \text{Pos}(\mathbf{m})), 
\end{equation}
where $\text{Pos}(\cdot)$ is the sinusoid positional embedding~\cite{transformer}. 
$\text{Mixup}((\mathbf{s}, \mathbf{x}), p^*)$ indicates the \emph{mixup sequence} of speech $\mathbf{s}$ and text $\mathbf{x}$ with probability $p^*$, which is fed into the \emph{translation encoder} for predicting the translation. 


\subsection{Self-learning with \method}
\label{sec:self-training}
With the help of our proposed \method, we are now able to access multimodal mixed sequences, in addition to the unimodal speech sequences. We integrate them into a self-learning framework. 
Specifically, we input both unimodal speech sequences and multimodal mixed sequences into the translation module (\emph{translation encoder} and \emph{translation decoder}).
In this way, translation of unimodal speech sequences focuses on the ST task itself, while the translation of multimodal mixed sequences is devoted to capture the connections between representations in different modalities. Besides, we try to regularize above two output predictions by minimizing the Jensen-Shannon Divergence (JSD) between two output distributions, which is
\begin{equation}
    \label{eq:kd}
    \begin{split}
        \mathcal{L}_\text{JSD}(\mathbf{s}, \mathbf{x}, \mathbf{y}, &p^*) = \sum_{i=1}^{|\mathbf{y}|}\text{JSD}\{p_\theta(\mathbf{y}_i|\mathbf{y}_{<i}, \mathbf{h}(\mathbf{s}))\| \\ &p_\theta(\mathbf{y}_i|\mathbf{y}_{<i}, \mathbf{h}(\text{Mixup}((\mathbf{s}, \mathbf{x}), p^*)))\},
    \end{split}
\end{equation}
where $\mathbf{h}(\cdot)$ is the contextual representation outputted by the \emph{translation encoder}. $p_\theta(\mathbf{y}_i|\mathbf{y}_{<i}, \mathbf{h}(\mathbf{s}))$ is the predicted probability distribution of the $i$-th target token given the speech sequence $\mathbf{s}$ as input, and $p_\theta(\mathbf{y}_i|\mathbf{y}_{<i}, \mathbf{h}(\text{Mixup}((\mathbf{s}, \mathbf{x}), p^*)))$ is that given the multimodal mixed sequence as input.

With the cross-entropy losses of two forward passes, the final training objective is as follows:
\begin{equation}
    \label{eq:final-loss}
    \begin{split}
        \mathcal{L} = &\ \mathcal{L}_{\text{CE}}(\mathbf{s}, \mathbf{y}) + \mathcal{L}_{\text{CE}}(\text{Mixup}((\mathbf{s}, \mathbf{x}), p^*), \mathbf{y}) \\
        &+\lambda\mathcal{L}_\text{JSD}(\mathbf{s}, \mathbf{x}, \mathbf{y}, p^*),
    \end{split}
\end{equation}
where $\lambda$ is the coefficient weight to control $\mathcal{L}_\text{JSD}$. 

\subsection{Mixup Ratio Strategy}
\label{sec:mixup-strategy}
When using our proposed \method, an important question is how to determine the mixup ratio $p^*$. Here we try two strategies: \emph{static mixup ratio} and \emph{uncertainty-aware mixup ratio}.

\noindent\textbf{Static Mixup Ratio}~
We use the same mixup ratio $p^*$ for all instances throughout the whole training process. We will show how we determined this important hyper-parameter in Section \ref{sec:mixup-ratio}.

\noindent\textbf{Uncertainty-aware Mixup Ratio}~
With this strategy, we determine the mixup ratio for each instance according to the prediction uncertainty of the ST task, defined as the average entropy of predicted distributions of all target tokens:
\begin{equation}
    u=\frac{1}{|\mathbf{y}|}\sum_{i=1}^{|\mathbf{y}|}\text{Entropy}(p_\theta(\mathbf{y}_i|\mathbf{y}_{<i}, \mathbf{h}(\mathbf{s}))),
\end{equation}
and then we set the mixup ratio $p^*$ as follows:
\begin{equation}
    p^*=\sigma\left(\frac{u}{U}-\frac{1}{2}\right),
\end{equation}
where $U$ is a normalization factor which re-scales $u$ to $[0, 1]$, $\sigma(\cdot)$ is a sigmoid function to prevent $p^*$ from dropping too quickly. 

\section{Experiments}

\begin{table}[t]
    \centering
    \begin{tabular}{c|cr||cr}
        \toprule
         & \multicolumn{2}{c||}{\textbf{ST (MuST-C)}} & \multicolumn{2}{c}{\textbf{MT}} \\
         \textbf{En$\rightarrow$} & hours & \#sents & name & \#sents \\
        \midrule
         \textbf{De} & 408 & 234K & WMT16 & 4.6M \\
         \textbf{Fr} & 492 & 280K & WMT14 & 40.8M \\
         \textbf{Ru} & 489 & 270K & WMT16 & 2.5M \\
         \textbf{Es} & 504 & 270K & WMT13 & 15.2M \\
         \textbf{Ro} & 432 & 240K & WMT16 & 0.6M \\
         \textbf{It} & 465 & 258K & OPUS100 & 1.0M \\
         \textbf{Pt} & 385 & 211K & OPUS100 & 1.0M \\
         \textbf{Nl} & 442 & 253K & OPUS100 & 1.0M \\
        \bottomrule
    \end{tabular}
    \caption{Statistics of all datasets}
    \label{tab:data}
\end{table}

\begin{table*}[t]
\centering
\small
\resizebox{\textwidth}{!}{
\begin{tabular}{l|ccc|cccccccc|c}
\toprule
\multirow{2}{*}{\textbf{Models}} & \multicolumn{3}{c|}{\textbf{External   Data}} & \multicolumn{9}{c}{\textbf{BLEU}} \\
                        & Speech        & ASR       & MT       & En-De & En-Fr & En-Ru & En-Es & En-It & En-Ro & En-Pt & En-Nl & Avg. \\ 
\midrule
\multicolumn{13}{c}{\emph{Pretrain w/o external MT data}} \\
\midrule
Fairseq ST \citep{wang2020fairseqs2t}             & \texttimes             & \texttimes         & \texttimes        & 22.7   & 32.9   & 15.3 & 27.2 & 22.7 & 21.9 & 28.1 & 27.3 & 24.8  \\
AFS \citep{zhang-etal-2020-adaptive}                     & \texttimes             & \texttimes         & \texttimes        & 22.4   & 31.6   & 14.7 & 26.9 & 23.0 & 21.0 & 26.3 & 24.9 & 23.9 \\
DDT \citep{le-etal-2020-dual}                    & \texttimes             & \texttimes         & \texttimes        & 23.6   & 33.5   & 15.2 & 28.1 & 24.2 & 22.9 & 30.0 & 27.6 & 25.6 \\
Self-training \citep{pino2020self} & \checkmark & \checkmark & \texttimes & 25.2 & 34.5 & - & - & - & - & - & - & -\\
BiKD \citep{inaguma-etal-2021-source} & \texttimes & \texttimes & \texttimes & 25.3 & 35.3 & - & - & - & - & - & - & -\\
SATE \citep{xu-etal-2021-stacked}                   & \texttimes             & \texttimes         & \texttimes        & 25.2   & -      & -    & - & - & - & - & -  & -\\ 
XSTNet~\cite{ye2021end} & \checkmark & \texttimes & \texttimes  & 25.5 & 36.0 & 16.9 & 29.6 & 25.5 & \textbf{25.1} & \textbf{31.3} & 30.0 & \textbf{27.5} \\
W2V2-Transformer & \checkmark & \texttimes & \texttimes & 24.1 & 35.0 & 16.3 & 29.4 & 24.8 & 23.1 & 30.0 & 28.9 & 26.5 \\
\method & \checkmark & \texttimes & \texttimes & \textbf{25.6**} & \textbf{36.1**} & \textbf{17.1**} & \textbf{30.3**} & \textbf{25.6**} & 24.3** & 31.0** & \textbf{30.1**} & \textbf{27.5} \\
\midrule
\multicolumn{12}{c}{\emph{Pretrain w/ external MT data}} \\
\midrule
MTL \citep{generalmtl}                    & \texttimes             & \texttimes         & \checkmark        & 23.9   & 33.1   & -  & 28.6 & - & - & - & - & -   \\
FAT-ST \citep{FusAT}                 & \checkmark             & \checkmark         & \checkmark        & 25.5   & -      & -    & 30.8 & - & - & - & 30.1 & -  \\
JT-S-MT \citep{tang-etal-2021-improving}                & \texttimes             & \texttimes         & \checkmark        & 26.8   & 37.4   & - & \textbf{31.0} & - & - & - & - & -     \\
SATE \citep{xu-etal-2021-stacked}                   & \texttimes             & \checkmark         & \checkmark        & 28.1\textsuperscript{\textdagger}   & -      & -    & - & - & - & - & - & -  \\ 
Chimera \citep{han-etal-2021-learning}                & \checkmark             & \texttimes         & \checkmark        & 27.1\textsuperscript{\textdagger}   & 35.6   & 17.4 & 30.6 & 25.0 & 24.0 & 30.2 & 29.2 & 27.4 \\
XSTNet~\cite{ye2021end} & \checkmark & \texttimes & \checkmark  & 27.8 & \textbf{38.0} & \textbf{18.5} & 30.8 & \textbf{26.4} & \textbf{25.7} & \textbf{32.4} & \textbf{31.2} & \textbf{28.8} \\
W2V2-Transformer        & \checkmark             & \texttimes         & \checkmark        & 26.9   &   36.6     &  17.3 & 30.0 & 25.4 & 23.9 & 30.7 & 29.6 & 27.6  \\
\method                & \checkmark             & \texttimes         & \checkmark        & \textbf{28.7**}   &    37.4**    &   17.8** & \textbf{31.0**} & 25.8* & 24.5** & 31.7** & 30.5** & 28.4     \\ 
\bottomrule
\end{tabular}}
\caption{BLEU scores on MuST-C \texttt{tst-COMMON} set. "Speech" denotes unlabeled audio data. \textdagger \ use OpenSubtitles \citep{lison-tiedemann-2016-opensubtitles2016} as external MT data. * and ** mean the improvements over W2V2-Transformer baseline is statistically
significant ($p < 0.05$ and $p < 0.01$, respectively).}
\label{tab:main}
\end{table*}

\begin{table}[t]
    \centering
    \small
    \resizebox{\linewidth}{!}{
    \begin{tabular}{l|ccc}
    \toprule
    \textbf{Models} & \textbf{WER}↓ & \textbf{MT BLEU}↑ & \textbf{ST BLEU}↑ \\
    \midrule
    Cascaded & 9.9 & 31.7 & 27.5 \\
    W2V2-Transformer & - & 31.7 & 26.9 \\
    \method & - & 31.7 & \textbf{28.7**} \\
    \bottomrule
    \end{tabular}}
    \caption{Comparison with cascaded baseline on MuST-C En-De \texttt{tst-COMMON} set. ** mean the improvements over cascaded baseline is statistically significant ($p < 0.01$).}
    \label{tab:cascaded}
\end{table}

\subsection{Datasets}
\noindent\textbf{MuST-C}~
We conduct experiments on MuST-C \citep{di-gangi-etal-2019-must} dataset. MuST-C is a multilingual speech translation dataset, which contains translations from English (En) to 8 languages: German (De), French (Fr), Russian (Ru), Spanish (Es), Italian (It), Romanian (Ro), Portuguese (Pt), and Dutch (Nl). It is one of the largest speech translation datasets currently, which contains at least 385 hours of audio recordings from TED Talks, with their manual transcriptions and translations at the sentence level. We use \texttt{dev} set for validation and \texttt{tst-COMMON} set for test. 

\noindent\textbf{MT Datasets}~
Our model architecture allows us to utilize external parallel sentence pairs in large-scale machine translation datasets. Therefore, we incorporate data from WMT for En-De, En-Fr, En-Ru, En-Es, En-Ro, and OPUS100\footnote{\url{http://opus.nlpl.eu/opus-100.php}} for En-Pt, En-It, En-Nl, as pretraining corpora. The detailed statistics of all datasets included are shown in Table~\ref{tab:data}.

\subsection{Experimental setups}

\noindent\textbf{Pre-processing}~
For speech input, we use the raw 16-bit 16kHz mono-channel audio wave. To perform word-level force alignment, we use Montreal Forced Aligner\footnote{\url{https://github.com/MontrealCorpusTools/Montreal-Forced-Aligner}} toolkit, whose acoustic model is trained with LibriSpeech~\citep{librispeech}. 
For text input, we remove the punctuation from the source texts for the ST dataset. Both source and target texts are case-sensitive. 
For each translation direction, we use a unigram SentencePiece\footnote{\url{https://github.com/google/sentencepiece}} model to learn a vocabulary on the text data from ST dataset, and use it to segment text from both ST and MT corpora into subword units. The vocabulary is shared for source and target with a size of 10k.

\noindent\textbf{Model Configuration}~
Our model consists of three modules. For the \emph{acoustic encoder}, we use Wav2vec2.0~\citep{baevski2020wav2vec} following the base configuration, which is pretrained on audio data from LibriSpeech \citep{librispeech} without finetuning\footnote{Model can be downloaded at \url{https://dl.fbaipublicfiles.com/fairseq/wav2vec/wav2vec_small.pt}}. 
We add two additional 1-dimensional convolutional layers to further shrink the audio, with kernel size 5, stride size 2, padding 2, and hidden dimension 1024. For the \emph{translation encoder}, we use $N_e=6$ transformer encoder layers. For the \emph{translation decoder}, we use $N_d=6$ transformer decoder layers. Each of these transformer layers comprises 512 hidden units, 8 attention heads, and 2048 feed-forward hidden units. 

\noindent\textbf{Training and Inference}~
We train our model in a pretrain-finetune manner. During pretraining, we train the MT model i.e., \emph{translation encoder} and \emph{translation decoder}, with \emph{transcription-translation} pairs. The learning rate is 7e-4. We train the model with at most 33k input tokens per batch. During finetuning, the learning rate is set to 1e-4. We finetune the whole model up to 25 epochs to avoid overfitting, with at most 16M source audio frames per batch. The training will early-stop if the loss on \emph{dev} set did not decrease for ten epochs. During both pretraining and finetuning, we use an Adam optimizer \citep{adam} with $\beta_1=0.9, \beta_2=0.98$, and 4k warm-up updates. The learning rate will decrease proportionally to the inverse square root of the step number after warm-up. The dropout is set to 0.1, and the value of label smoothing is set to 0.1. We use the uncertainty-aware mixup ratio strategy by default, and the mixup ratio $p^*$ is set to 0.4 when using static strategy. The weight $\lambda$ of JSD loss is set to 1.0.

During inference, We average the checkpoints of the last 10 epochs for evaluation. We use beam search with a beam size of 5. 
We use sacreBLEU\footnote{\url{https://github.com/mjpost/sacrebleu}} \citep{post-2018-call} to compute case-sensitive detokenized BLEU \citep{papineni-etal-2002-bleu} scores and the statistical significance of translation results with paired bootstrap resampling \citep{koehn-2004-statistical} for a fair comparison\footnote{sacreBLEU signature: nrefs:1 | bs:1000 | seed:12345 | case:mixed | eff:no | tok:13a | smooth:exp | version:2.0.0}.
All models are trained on 8 Nvidia Tesla-V100 GPUs. We implement our models based on fairseq\footnote{\url{https://github.com/pytorch/fairseq}} \citep{ott2019fairseq}.

\paragraph{Baseline Systems}

We compare our method with several strong end-to-end ST systems including: Fairseq ST \citep{wang2020fairseqs2t}, AFS \citep{zhang-etal-2020-adaptive}, DDT \citep{le-etal-2020-dual}, MTL \citep{generalmtl}, Self-training \citep{pino2020self}, BiKD \citep{inaguma-etal-2021-source}, FAT-ST \citep{FusAT}, JT-S-MT \citep{tang-etal-2021-improving}, SATE \citep{xu-etal-2021-stacked}, Chimera \citep{han-etal-2021-learning} and XSTNet~\citep{ye2021end}. Besides, we implement a strong baseline W2V2-Transformer based on Wav2vec2.0. It has the same model architecture as our proposed \method and is pretrained in the same way. The only difference is that it is only finetuned on the ST task, while we adopt a self-learning framework during finetuning.

\section{Results and Analysis}

\subsection{Results on MuST-C Dataset}

\noindent\textbf{Comparison with End-to-end Baselines}~
As shown in Table \ref{tab:main}, our implemented W2V2-Transformer is a relatively strong baseline, which proves the effectiveness of Wav2vec2.0 module and MT pretraining. 
Without external MT data, our method achieves an improvement of 1.0 BLEU (average over 8 directions) over the strong baseline, which proves our proposed self-learning framework could effectively improve the performance of the ST task. It even outperforms baselines with external MT data on En-Es, En-It, En-Ro, En-Pt, and En-Nl. When we introduce additional MT data, our method also yields a 0.8 BLEU improvement compared with baseline. 
Note that our performance is slightly worse than XSTNet~\citep{ye2021end}. However, our method is orthogonal with theirs, which focuses on the training procedure of end-to-end ST model. We will investigate how to combine them together in the future.

\noindent\textbf{Comparison with Cascaded Baseline}~
We also implement a strong cascaded system, whose ASR part is composed of a pretrained Wav2vec2.0 module and 6 transformer decoder layers, and the MT part is the same as our pretrained MT module. 
Both cascaded systems and end-to-end models are trained with the same data ($\mathcal{D}$ and $\mathcal{D}_\text{MT}$). 
As shown in Table \ref{tab:cascaded}, the end-to-end baseline W2V2-Transformer is inferior to the cascaded system, but our method significantly outperforms it, which shows the potential of our \method method.

\subsection{Ablation Studies}
\begin{table}[t]
\centering
\small
\begin{tabular}{ccc|c}
\toprule
    \textbf{Mixup Ratio} & \textbf{\method Trans.} & \textbf{JSD}  & \textbf{BLEU} \\
    \midrule
    uncertainty-aware & \checkmark & \checkmark &  \textbf{28.7**} \\
    static & \checkmark & \checkmark &  28.5** \\
    static & \checkmark & \texttimes &  27.9** \\
    static & \texttimes & \texttimes &  26.9 \\
\bottomrule
\end{tabular}
\caption{BLEU scores on MuST-C En-De \texttt{tst-COMMON} set with different auxiliary training objectives. \method Trans. indicates the criterion entropy loss of translation of multimodal mixed sequence $\mathcal{L}_\text{CE}(\text{Mixup}((\mathbf{s}, \mathbf{x}), p^*), \mathbf{y})$. ** mean the improvements over W2V2-Transformer baseline (last row in the table) is statistically
significant ($p < 0.01$).}
\label{tab:ablation-objectives}
\end{table}
\noindent\textbf{Is Each Learning Objective Effective?}~
As shown in Equation~\ref{eq:final-loss}, our training objective contains three terms. Besides the cross-entropy objective $\mathcal{L}_\text{CE}(\mathbf{s}, \mathbf{y})$ for speech translation, we investigate the effects of the other two auxiliary training objectives. 
As shown in Table \ref{tab:ablation-objectives}, when we input the additional multimodal mixed sequence into the model and optimize the cross-entropy loss (Line 3), it can already outperform the baseline (Line 4) significantly. When we regularize two output predictions with JSD loss (Line 2), the performance can be further boosted.


\noindent\textbf{The uncertainty-aware strategy reduces the cost for searching mixup ratio and has  better performance.}
We present two different mixup ratio strategies in Section~\ref{sec:mixup-strategy}. To evaluate their impacts, we conduct another ablation study on MuST-C En-De. We observe that the BLEU scores on \texttt{tst-COMMON} set are 28.5 and 28.7 for \emph{static strategy} and \emph{uncertainty-aware strategy}, respectively. 
The \emph{uncertainty-aware strategy} can slightly improve the performance, and more importantly, it lowers the manual cost for searching an optimal mixup ratio to get the best performance.

\subsection{What is the Optimal Mixup Ratio?}
\label{sec:mixup-ratio}
\begin{figure}[t]
    \centering
    \includegraphics[width=\linewidth]{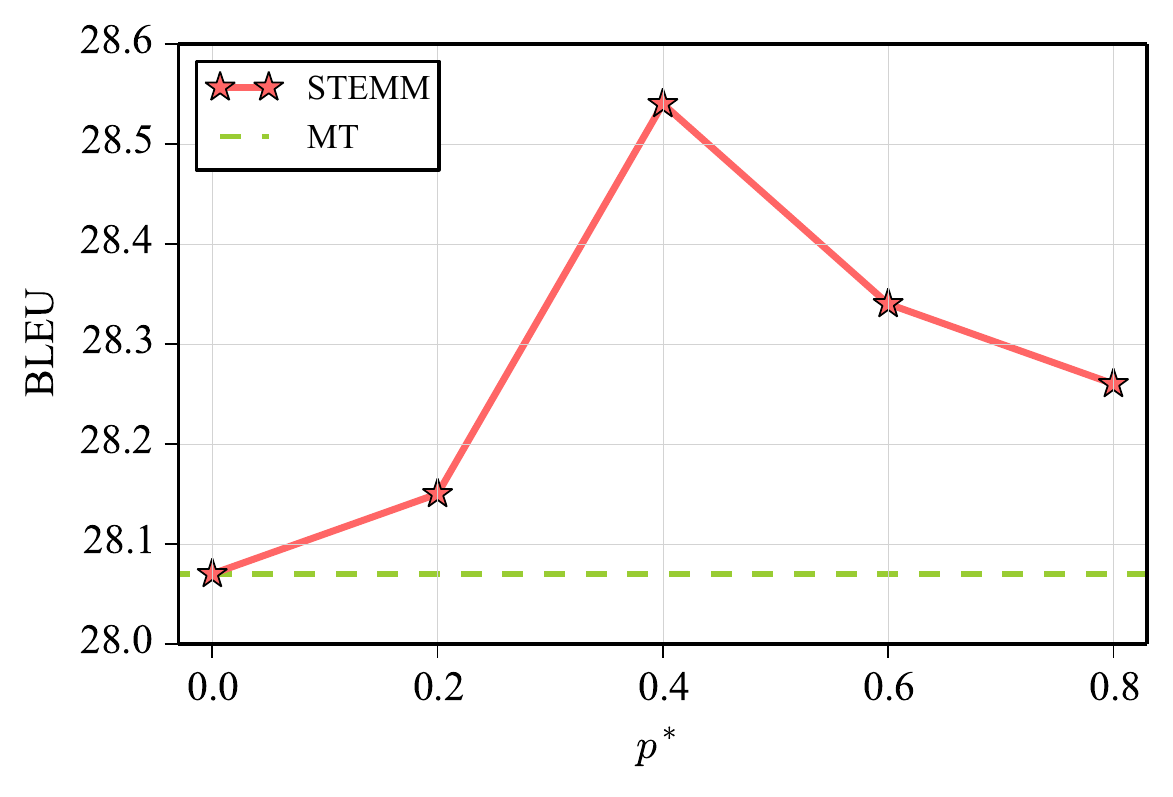}
    \caption{BLEU scores on MuST-C En-De \texttt{tst-COMMON} set with different mixup ratio $p^*$. Our method achieves best performance when $p^*=0.4$. When $p^*=0.0$, \method will degrade to text-only sequence, which we denote as MT.}
    \label{fig:mixup-ratio}
\end{figure}

When using \emph{static mixup ratio strategy}, it is important to choose the mixup ratio $p^*$. We constrain $p^*$ in $[0.0, 0.2, 0.4, 0.6, 0.8]$ for experiments on MuST-C En-De \texttt{tst-COMMON} set, as shown in Figure \ref{fig:mixup-ratio}. 
When $p^*=0.0$, the translation task with the mixed sequence as input degrades to the MT task.
We interestingly find that self-learning with MT tasks performed the worst (i.e. lowest BLEU) than self-learning with \method at other mixup ratios.
This confirms what we mentioned in Section~\ref{sec:intro}, that the representation discrepancy between speech and text makes the MT task an inferior boost to ST.

\begin{figure}[t]
    \centering
    \includegraphics[width=\linewidth]{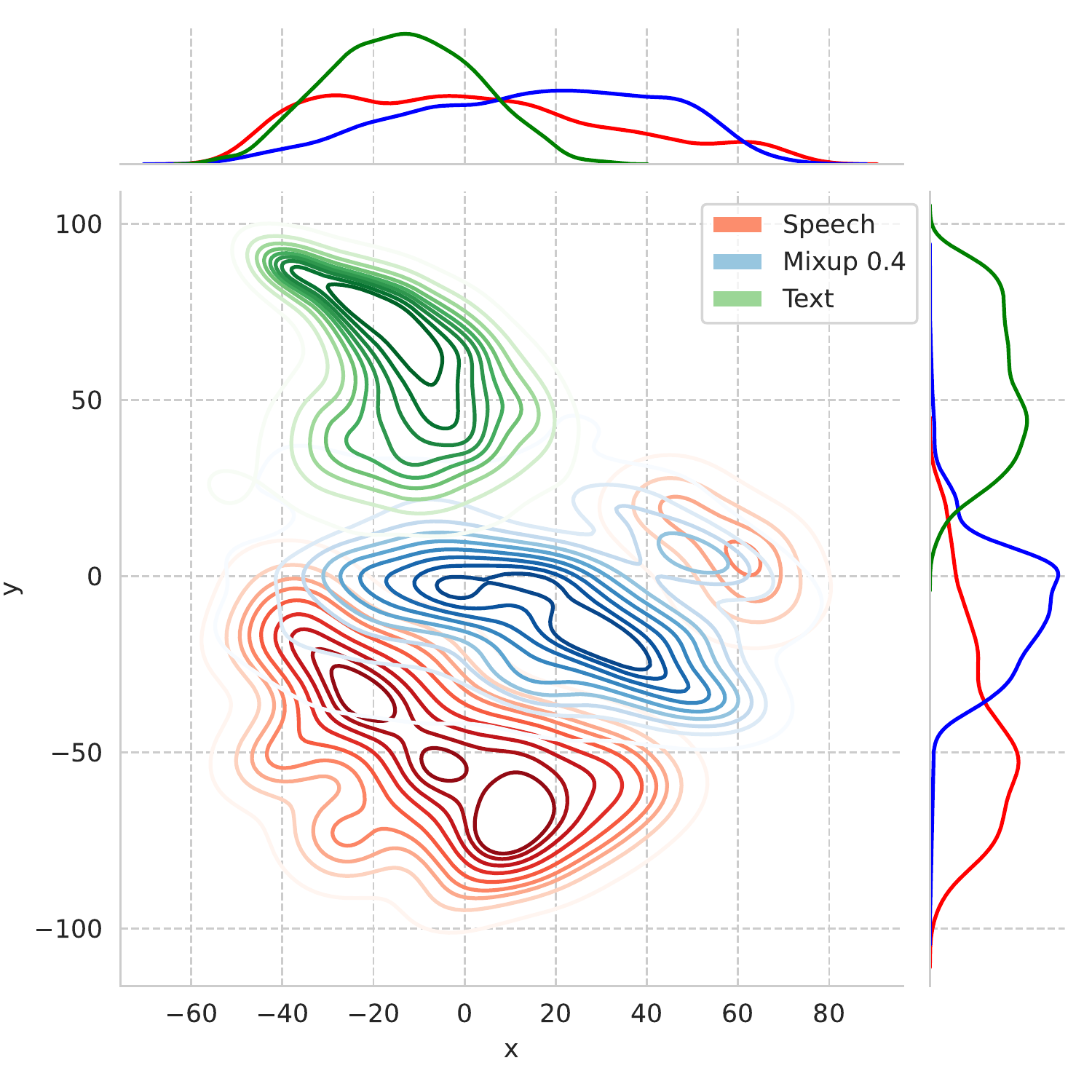}
    \caption{The bivariate kernel density estimation visualization of the averaged sentence representation of the speech, text and \method sequences after pretraining. 
    T-SNE algorithm is applied to reduce the 512-dim representations to two dimensions. The green line stands for the averaged sentence embedding. The red line stands for the averaged speech representation. the blue line is the representation for \method with mixup ratio $p^*=0.4$.
    We observe that the representation of the mixed sequence is in between that of speech and text, which fills the gap between the representation of speech sequences and text sequences. Best view in color.}
    \label{fig:init-reps}
\end{figure}

Our method achieves the best performance at $p^*=0.4$.
To find a reasonable explanation, we do a more in-depth study of the representation of the speech, text, and their mixup sequence (\method). In Figure~\ref{fig:init-reps}, we take out the sequential representation of the speech (output of acoustic encoder), text sequences (output of embedding layer), and the \method sequences, average them over the sequence dimension, and apply the T-SNE dimensionality reduction algorithm to reduce the 512 dimensions to two dimensions. We plot the bivariate kernel density estimation based on the reduced 2-dim representation. 
We find that when $p^*=0.4$, the mixup representation just lies between the representation of speech and text sequences. That is why it calibrates the cross-modal representation discrepancy more easily and gets the best ST performance.

\subsection{Can Our Model Alleviate Cross-modal Representation Discrepancy?}
To examine whether our method alleviates the cross-modal representation discrepancy, we conduct some analysis of cross-modal word representations. As described in Section~\ref{sec:mixup}, for each word unit $w_i$, we identify the corresponding segment of speech representation $[\mathbf{a}_{l_i} : \mathbf{a}_{r_i}]$ and text embedding $[\mathbf{e}_{m_i} : \mathbf{e}_{n_i}]$. We define the word representation in each modality as follows:
\begin{gather}
    \bm{\alpha}_i=\text{AvgPool}([\mathbf{a}_{l_i} : \mathbf{a}_{r_i}]), \\
    \bm{\varepsilon}_i=\text{AvgPool}([\mathbf{e}_{m_i} : \mathbf{e}_{n_i}]),
\end{gather}
where $\text{AvgPool()}$ denotes average-pooling operation across the sequence dimension, $\bm{\alpha}_i$ and $\bm{\varepsilon}_i$ denote the representation of word unit $w_i$ in speech and text modalities, respectively.

\begin{figure}[t]
    \centering
    \includegraphics[width=\linewidth]{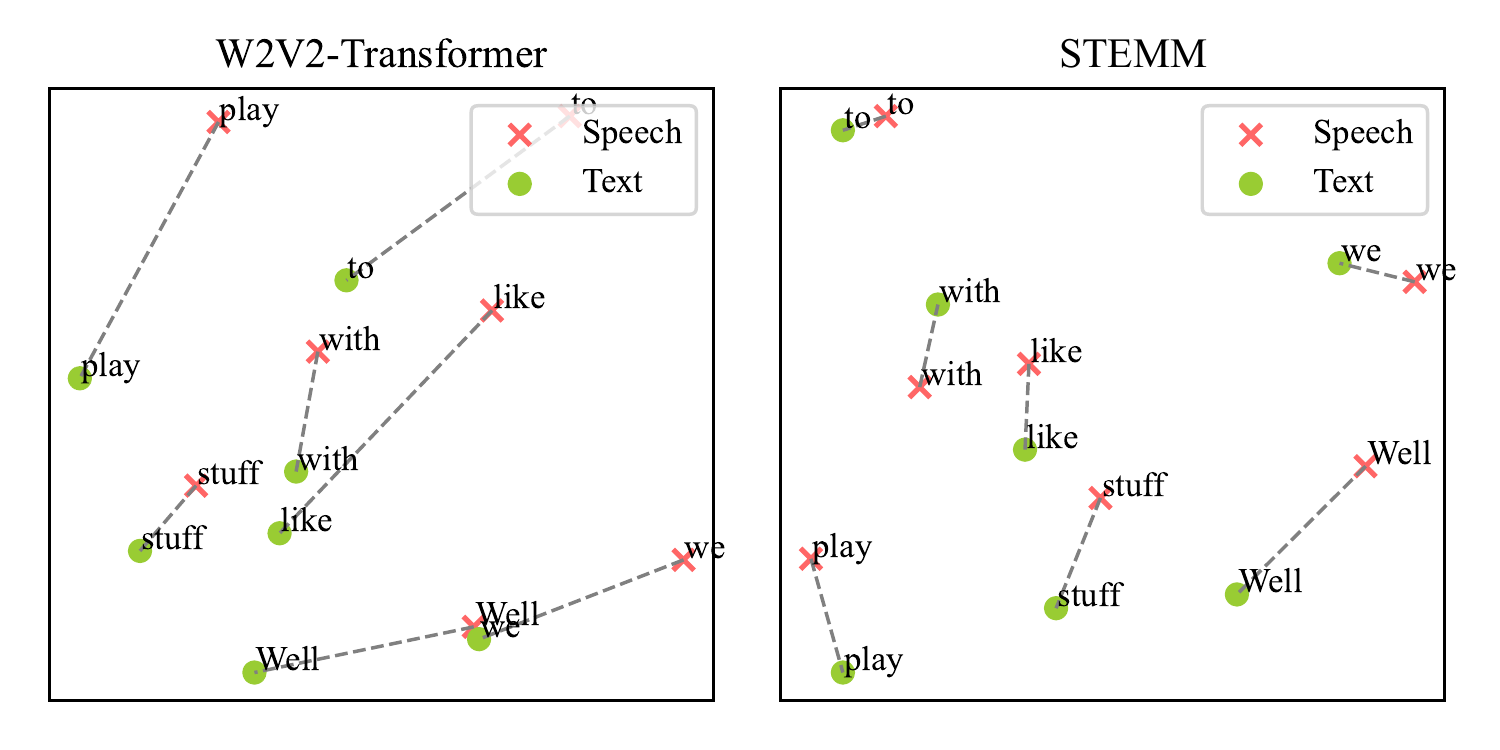}
    \caption{Visualization of word representations in speech and text modalities. We visualize the representations by reducing the dimension with Principal Component Analysis (PCA). Our method brings word representation within different modalities closer. }
    \label{fig:word-reps}
\end{figure}

We calculate the average cosine similarity between $\bm{\alpha}_i$ and $\bm{\varepsilon}_i$ over all word units $w_i$ in MuST-C En-De \texttt{tst-COMMON} set. As shown in Table~\ref{tab:word_sim}, our method could significantly improve the similarity of word representations across modalities over baseline. We believe it is because when training with our proposed \method, the speech segment and text segment of a word will appear in a similar multimodal context, which leads to similar representations. We also show the visualization of an example in Figure~\ref{fig:word-reps}, we can observe that our method brings word representations within different modalities closer compared with baseline.

\begin{table}[t]
    \centering
    \begin{tabular}{l|c}
    \toprule
    \textbf{Models} & \textbf{Similarity (\%)}  \\
    \midrule
    W2V2-Transformer & 32.31 \\
    \method & \textbf{51.89} \\
    \bottomrule
    \end{tabular}
    \caption{Comparison of word-level representation similarity across modalities.}
    \label{tab:word_sim}
\end{table}

\subsection{How the Size of MT Data Influences Performance?}
\begin{figure}[t]
    \centering
    \includegraphics[width=\linewidth]{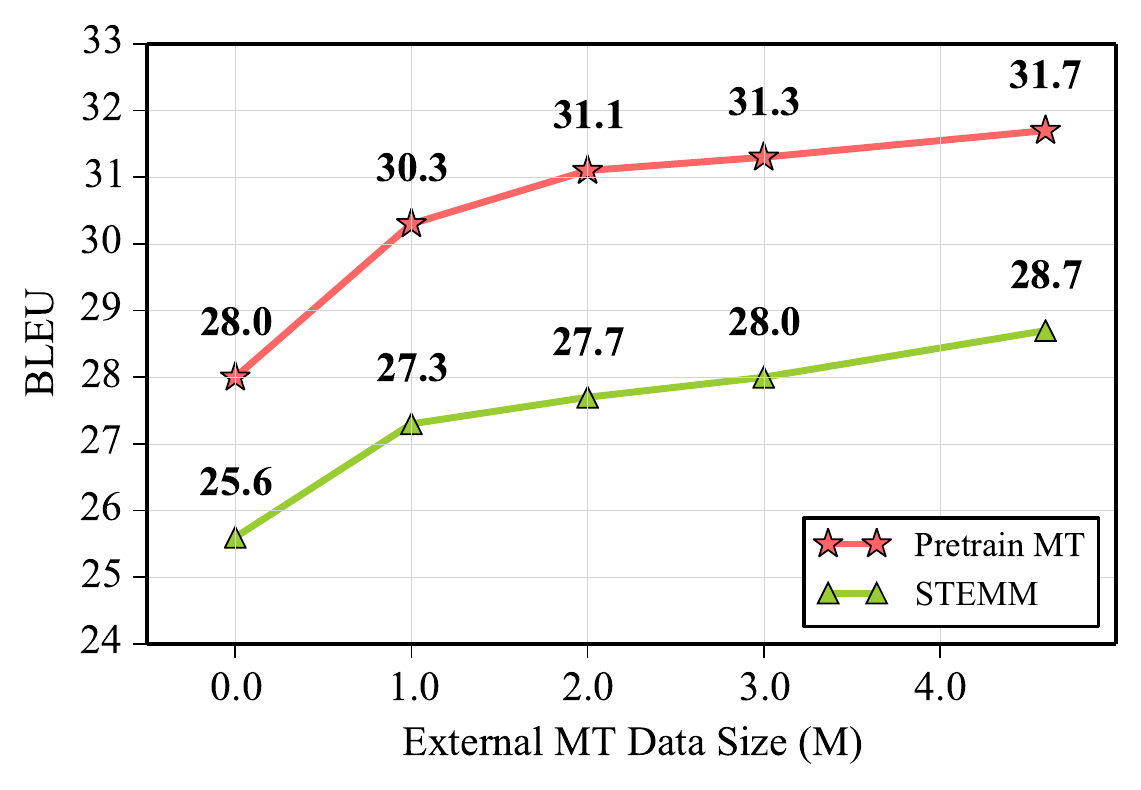}
    \caption{Curve of BLEU scores on MuST-C En-De \texttt{tst-COMMON} against the size of external MT data used during pretraining.}
    \label{fig:mt-amount}
\end{figure}

One important contributor to our excellent performance is the usage of external MT data. Therefore, how the amount of MT data affects the final performance is an important question. We vary the amount of available external MT data during pretraining on En-De direction. As shown in Figure~\ref{fig:mt-amount}, we observe a continuous improvement of BLEU scores with the increase of MT data, which shows that external MT data is helpful to improve ST.

\subsection{Can the Final Model still Perform MT Task?}

Our model is first pretrained on the MT task and then finetune for ST. An important question is whether there is a catastrophic forgetting problem during finetuning. We evaluate the model on the MT task and show the result in Table \ref{tab:mt-performance}. We observe that when we only finetune the model on the ST task (W2V2-Transformer), the ability of text translation will be forgotten a lot. In contrast, when we use our self-learning framework during finetuning, even though there is no MT task, the MT capability can still be preserved.

\begin{table}[t]
    \centering
    \begin{tabular}{l|c}
    \toprule
    \textbf{Models} & \textbf{BLEU} \\
    \midrule
    Pretrained MT & \textbf{31.7}  \\
    W2V2-Transformer & 19.5 \\
    \method & 31.5 \\
    \bottomrule
    \end{tabular}
    \caption{BLEU scores of MT task on MuST-C En-De \texttt{tst-COMMON} set. Our proposed method almost preserves the text translation capability of pretrained MT model.}
    \label{tab:mt-performance}
\end{table}




\section{Related Works}
\noindent\textbf{End-to-end ST}~
To overcome the error propagation and high latency in the cascaded ST systems, \citet{berard2016listen, duong2016attentional} proved the potential of end-to-end ST without intermediate transcription, which has attracted much attention in recent years \citep{vila2018end, salesky2018towards, salesky2019fluent, digangi2019adapting, digangi2019enhancing, bahar2019comparative, inaguma2020espnet}. Since it is difficult to train an end-to-end ST model directly, some training techniques like pretraining \citep{weiss2017sequence, berard2018end, bansal2019pre, stoian2020analyzing, wang2020bridging, pino2020self, dong2021consecutive, alinejad2020effectively,zheng2021fused, xu-etal-2021-stacked}, multi-task learning \citep{le-etal-2020-dual, vydana2021jointly, generalmtl, ye2021end, tang-etal-2021-improving}, curriculum learning \citep{kano2017structured, wang-etal-2020-curriculum}, and meta-learning \citep{indurthi2020data} have been applied. To overcome the scarcity of ST data, \citet{jia2019leveraging, pino2019harnessing, bahar2019using} proposed to generate synthesized data based on ASR and MT corpora. 
To overcome the modality gap, \citet{han-etal-2021-learning, huang-etal-2021-adast, xu-etal-2021-stacked} further encode acoustic states which are more adaptive to the decoder. 
Previous works have mentioned that the modality gap between speech and text is one of the obstacles in the speech translation task, and to overcome such gap, 
one branch of the works~\cite{liu2020bridging, dong2021listen, xu-etal-2021-stacked} introduced a second encoder based on the conventional encoder-decoder model, to extract semantic information of speech and text. Recently, \citet{han-etal-2021-learning} built a shared semantic projection module that simulates the human brain,
while in this work, we explored how to construct an intermediate state of the two modalities via the recent mixup method (\textit{i.e.} \textbf{S}peech-\textbf{TE}xt \textbf{M}anifold \textbf{M}ixup) to narrow such gap. Note that our work is orthogonal with \citet{ye2021end}'s study in training procedure of end-to-end ST model.

\noindent\textbf{Mixup}~
Our work is inspired by the mixup strategy.
\citet{zhang2018mixup} first proposed mixup as a data augmentation method to improve the robustness and the generalization of the model, where additional data are constructed as the linear interpolation of two random examples and their labels at the surface level. \citet{verma2019manifold} extended the surface-level mixup to the hidden representation by constructing \textit{manifold mixup} interpolations.
Recent work has introduced mixup on machine translation~\cite{zhang-etal-2019-bridging, li-etal-2021-mixup-decoding, guo-etal-2022-prediction-difference, fang-etal-2022-PLUVR}, sentence classification~\cite{chen2020mixtext, jindal2020augmenting,sun2020mixup}, multilingual understanding~\cite{yang2022enhancing}, and speech recognition~\cite{medennikov2018investigation, Sun2021SemanticDA, DBLP:journals/corr/abs-2104-01393, meng2021mixspeech}, and obtained enhancements.
Our approach is the first to introduce the idea of manifold mixup to the speech translation task with two modalities, speech, and text.

\section{Conclusion}
In this paper, we propose a \textbf{S}peech-\textbf{TE}xt \textbf{M}anifold \textbf{M}ixup (\method) method to mix up the speech representation sequences and word embedding sequences. Based on \method, we adopt a self-learning framework, which learns the translation of unimodal speech sequences and multimodal mixed sequences in parallel, and regularizes their output predictions. Experiments and analysis demonstrate the effectiveness of our proposed method, which can alleviate the cross-modal representation discrepancy to some extent and improve the performance of ST. In the future, we will explore how to further eliminate this discrepancy and fill the cross-modal transfer gap for ST.

\section*{Acknowledgements}
We thank all the anonymous reviewers for their insightful and valuable comments. This work was supported by National Key R\&D Program of China (NO. 2017YFE0192900).

\bibliography{anthology,custom}
\bibliographystyle{acl_natbib}

\newpage
\appendix

\end{document}